\documentclass[11pt]{article}
\usepackage{coling2020}
\usepackage{times}
\usepackage{url}
\usepackage{latexsym}

\usepackage{microtype}

\usepackage{tcolorbox}

\usepackage{CJKutf8}
\usepackage{amsfonts}
\usepackage{amsmath,amssymb,bm}

\usepackage{array}
\usepackage{multirow}
\usepackage{makecell}
\usepackage{threeparttable}
\usepackage{booktabs}
\usepackage{mathtools}
\usepackage{color}
\usepackage{graphicx}
\usepackage{subfigure}

\title{End to End Chinese Lexical Fusion Recognition with Sememe Knowledge}

\author{Yijiang Liu \\
  School of Cyber \\ 
  Science and Engineering\\
  Wuhan University, China \\
  {\tt cslyj@whu.edu.cn} \\\And
  Meishan Zhang \\
  School of New Media \\
  and Communication\\
  Tianjin University, China\\
  {\tt mason.zms@gmail.com} \\\And
  Donghong Ji \\
  School of Cyber \\ 
  Science and Engineering\\
  Wuhan University, China \\
  {\tt dhji@whu.edu.cn} \\
  }

\date{}

\begin{document}
\begin{CJK}{UTF8}{gbsn}
\maketitle
\begin{abstract}
In this paper, we present Chinese lexical fusion recognition,
a new task which could be regarded as one kind of coreference recognition.
First, we introduce the task in detail,
showing the relationship with coreference recognition and differences from the existing tasks.
Second, we propose an end-to-end model for the task,
handling mentions as well as coreference relationship jointly.
The model exploits the state-of-the-art contextualized BERT representations as encoder,
and is further enhanced with the sememe knowledge from HowNet by graph attention networks.
We manually annotate a benchmark dataset for the task and then conduct experiments on it.
Results demonstrate that our final model is effective and competitive for the task.
Detailed analysis is offered for comprehensively understanding the new task and our proposed model.
\end{abstract}

\section{Introduction}
Coreference is one important topic in linguistics \cite{gordon1998representation,pinillos2011coreference},
and coreference recognition has been extensively researched in the natural language processing (NLP) community \cite{ng-cardie-2002-improving,lee-etal-2017-end,qi2012improve,fei-etal-2019-end}.
There are different kinds of coreferences, such as pronoun anaphora and abbreviation \cite{mitkov-1998-robust,Ruslan1999Anaphora,mu2002definite,choubey-huang-2018-improving}.
Here we examine the phenomenon of Chinese lexical fusion,
where two closely related words in a paragraph are united by
a fusion form of the same meaning,
and the fusion word can be seen as the coreference of the two separation words.
Since the fusion words are always out-of-vocabulary (OOV) words in downstream paragraph-level tasks such as reading comprehension, summarization, machine translation and etc.,
which can hinder the overall understanding and lead to degraded performance,
the new task can offer informative knowledge to these tasks.

\newtcbox{\fusionbox}[1][red]
  {on line,
    colback = #1!50!white, colframe = #1!50!black,
    boxsep = 0pt, left = 1pt, right = 1pt, top = 2pt, bottom = 2pt,
    boxrule = 0pt, bottomrule = 0pt, toprule = 0pt}

\newtcbox{\separatbox}[1][red]
  {on line,
    colback = #1!20!white, colframe = #1!50!black,
    boxsep = 0pt, left = 1pt, right = 1pt, top = 2pt, bottom = 2pt,
    boxrule = 0pt, bottomrule = 0pt, toprule = 0pt}

\begin{table}[h]
\centering
\begin{tabular}{p{15cm}l}
\toprule
卡纳瓦罗打破\fusionbox[red]{受访}惯例，公开训练未\separatbox[blue]{接受}\separatbox[blue]{采访}。\\
Cannavaro broke the convention of \fusionbox[red]{accepting an} \fusionbox[red]{interview}, did not \separatbox[blue]{accept} an \separatbox[blue]{interview} for public training. \\
\midrule
首批医务人员昨日\fusionbox[red]{返杭}，其余人员预计两周内\separatbox[blue]{回到}\separatbox[blue]{杭州}。 \\
The first medical personnel \fusionbox[red]{returned to} \fusionbox[red]{Hangzhou} yesterday, the others are expected to \separatbox[blue]{return to} \separatbox[blue]{Hangzhou} within two weeks.\\
\midrule
央行宣布\separatbox[blue]{下调}人民币贷款基准\separatbox[blue]{利率}，这是央行近期进行的第二次\fusionbox[red]{降息}调整。 \\
The central bank announced to \separatbox[blue]{reduce} the benchmark \separatbox[blue]{interest rate} for RMB loans, which is the second time to \fusionbox[red]{reduce the interest rate} recently.\\
\bottomrule
\end{tabular}
\caption{\label{samples} Examples of Chinese lexical fusion.}
\end{table}

Table \ref{samples} shows three examples of Chinese lexical fusion.
In the examples,
``受访''(accept an interview), ``返杭''(returned to Hangzhou) and ``降息''(reduce the interest rate)
are the fusion words,
and their coreferences are ``接受(accept)/采访(interview)'',
``回到(return)/杭州(Hangzhou)'', and ``下调 (reduce)/利率(interest rate)'', respectively.
Each corresponding of the fusion word consists of two separation words.
Besides, each fusion word character is semantically corresponding to one of the separation words,
which can be regarded as fine-grained coreferences.
As shown, we have six fine-grained coreferences by the three example fusion words:
``受$\leftrightarrow$接受(accept)'', ``访$\leftrightarrow$采访 (interview)'',
``返$\leftrightarrow$回到(return)'',  ``杭$\leftrightarrow$杭州(Hangzhou)'',
``降$\leftrightarrow$下调 (reduce)'' and ``息$\leftrightarrow$利率(interest rate)''.

Lexical fusion is used frequently in the Chinese language.
Moreover, the fusion words are usually rarer words compared with their separation words coreferred,
which are more difficult to be handled by NLP models\cite{DBLP:conf/acl/ZhangY18,DBLP:conf/ijcai/GuiM0ZJH19}.
Luckily, the meaning of the fusion words can be derived from that of the separation words.
Thus recognizing the lexical fusion would be beneficial for downstream paragraph (or document)-level NLP applications
such as machine translation, information extraction, summarization, etc.\cite{li-yarowsky-2008-unsupervised,ferreira2013assessing,kundu-etal-2018-neural}.
For example, for deep semantic parsing or translation, the fusion words ``受访''(UNK) can be substituted directly by the separation words ``接受''(accept) and ``访问''(interview), as the same fusion words are rarely occurred in other paragraphs.

%

The recognition of lexical fusion can be accomplished by two subtasks.
Given one paragraph,
the fusion words, as well as the separation words should be detected as the first step,
which is referred to as mention detection.
Second, coreference recognition is performed over the detected mentions,
linking each character in the fusion words to their coreferences, respectively.
By the second step, full lexical fusion coreferences are also recognized concurrently.
The two steps can be conducted jointly in a single end-to-end model \cite{lee-etal-2017-end},
which helps avoid the error propagation problem,
and meanwhile, enable the two subtasks with full interaction.

In this paper, we present a competitive end-to-end model for lexical fusion recognition.
Contextual BERT representations \cite{devlin2019bert} are adopted as encoder inputs
as they have achieved great success in a number of NLP tasks \cite{8837002,zhou-etal-2019-bert,xu-etal-2019-bert}.
For mention detection, a CRF decoder \cite{huang2015bidirectional} is exploited to detect all mention words, including both the fusion and the separation words.
Further, we use a BiAffine decoder for coreference recognition \cite{zhang2016dependency,bekoulis2018joint},
determining a given mention pair either to be a coreference or not.

Since our task is semantic oriented,
we use the sememe knowledge provided in HowNet\cite{dong2003hownet}
to help capturing the semantic similarity
between the characters and the separation words.
HowNet has achieved success in many Chinese NLP tasks in recent years \cite{DBLP:conf/ijcai/DuanZX07,gu-etal-2018-language,ding-etal-2019-event,li-etal-2019-chinese}.
Both Chinese characters and words are defined by senses of sememe graphs in it,
and we exploit graph attention networks (GAT) \cite{velivckovic2017graph} to model the sememe graphs to enhance our encoder.


Finally, we manually construct a high-quality dataset to evaluate our models.
The dataset consists of 7,271 cases of the lexical fusion,
which are all exploited as the test instances.
To train our proposed models, we construct a pseudo dataset automatically from the web resource.
Experimental results show that the auto-constructed training dataset is highly effective for our task.
The end-to-end models achieved better performance than the pipeline models,
and meanwhile the sememe knowledge can also bring significant improvements for both the pipeline and end-to-end models.
Our final model can obtain an F-measure of 79.64\% for lexical fusion recognition.
Further, we conduct in-depth analysis work to understand the proposed task and our proposed models.

In summary, we make the following contributions in this paper:
\begin{itemize}
\item[(1)]We introduce a new task of lexical fusion recognition,
providing a gold-standard test dataset and an auto-constructed pseudo training dataset for the task, which can be used as a benchmark for future research.
\item[(2)]We propose a competitive end-to-end model for lexical fusion recognition task, which helps to integrate the mention recognition with coreference identification based on BERT representations.
\item[(3)]We make use of the sememe knowledge from HowNet to help capturing the semantic relevance between the characters in the fusion form and the separation words.
\end{itemize}
All the codes and datasets will be open-sourced at https://github.com/xxx
under Apache License 2.0.

\section{Chinese Lexical Fusion}
Lexical fusion refers to the phenomenon that a composition word semantically corresponds with related words nearby in the contexts.
The composition word can be seen as a compact form of the separation words in the paragraph,
where the composition word is referred to as the fusion form of the separation words.
The fusion form is often an OOV word,
which brings difficulties in understanding of a given paragraph.
We can use a tuple to define the phenomenon in a paragraph: $\langle w_\text{F}, w_{\text{A}1}, w_{\text{A}2}, \cdots  , w_{\text{A}n} \rangle (n \geq 2)$,
where $w_\text{F}$ denotes the fusion word, and $w_{\text{A}1}, w_{\text{A}2}, \cdots  , w_{\text{A}n}$ are the separation words.
We regard the $w_\text{F}$ as a coreference of $w_{\text{A}1}, w_{\text{A}2}, \cdots  , w_{\text{A}n}$.
More detailedly, the number $n$ equals to the length of word $w_\text{F}$,
and the $i$th character of word $w_\text{F}$ corresponds to the separation word $w_{\text{A}i}$ as one fine-grained coreference.
The task is highly semantic-oriented, as all the fine-grained coreferences are semantically related.

Notice that Chinese lexical fusion has the following attributes,
making it different from others:
\begin{itemize}
\item[(1)] Each character of word $w_\text{F}$ must correspond to one and only one separation word.
The one-one mappings (clusters) should be strictly abided by.
For example,
$\langle$二圣 (two saints), 李治 (Zhi Li), 武则天 (Zetian Wu) $\rangle$ \footnote{The last two words of the tuple refer to names of two distinguished person. }
is not a lexical fusion phenomenon, because both the two characters ``二'' (two) and ``圣'' (saint) can not associate with any of the separations.
\item[(2)] For each fine-grained coreference, the $i$th character of word $w_\text{F}$ is mostly borrowed from its separated coreference $w_{\text{A}i}$ directly,
but it is not always true.
A semantically-equaled character could be used as well, which greatly increases the difficulty of Chinese lexical fusion recognition.
The given examples in Table \ref{samples} demonstrate the rule. As shown, the coreferences ``返$\leftrightarrow$回到(return)'', ``降$\leftrightarrow$下调 (reduce)'' and ``息$\leftrightarrow$利率(interest rate)'' are all of these cases.
\item[(3)] The separation words $w_{\text{A}1}, w_{\text{A}2}, \cdots  , w_{\text{A}n}$ are semantically independent, especially not indicating a single specific entity.
Thus the fine-grained coreferences are semantically different in lexical fusion.
For example, $\langle$北大 (Peking University), 北京 (Peking), 大学 (University) $\rangle$,
the semantic meaning of ``北京 (Peking) 大学 (University)'' is inseparable.
\item[(4)] The Chinese lexical fusion phenomenon could be either forward or backward.
Table \ref{samples} shows two examples of backward references (i.e., ``受访$\leftrightarrow$接受/采访 (accept an interview)'' and ``返杭$\leftrightarrow$回到/杭州 (returned to Hangzhou)'')
and one example of forward reference (i.e., ``降息$\leftrightarrow$下调/利率 (reduce the interest rate)'').
\end{itemize}
By the above characters, we can easily differentiate the Chinese lexical fusion with several other closely-related linguistic phenomena,
as an example, the abbreviation, the illustrated negative examples in item (1) and (3) are both abbreviations.

For simplicity, we limit our focus on $n=2$ in this work, because this situation occupies over 99\% of all the lexical fusion cases
according to our preliminary statistics.
Thus we can call tuples as triples all though this paper.

\section{Method}
We built an end-to-end model for the lexical fusion recognition task.
The recognition of lexical fusion consists of two steps.
First mention detection is performed to collect all mention words,
including the fusion words and their separations.
By applying the BIO tagging scheme associating with mention types, 
this subtask can be converted into a typical sequence labeling task \cite{ohta2012open,DBLP:conf/acl/MaH16}.
Second, we conduct pair-wise character-word clustering,
obtaining one-one fine-grained coreferences and reformatting them to triples of $\langle w_\text{F}, w_{\text{A}1}, w_{\text{A}2} \rangle$.
The end-to-end model alleviates the problem of error propagation in pipeline methods through a joint way.
The model consists of an encoder, a CRF decoder for mention recognition and a biaffine decoder for coreference recognition.
Further, we enhance the encoder with sememe knowledge from HowNet.

%
%

\subsection{Basic Model}
\paragraph{Encoder}
We exploit the BERT as the basic encoder because it has achieved state-of-the-art performances for a number of NLP tasks \cite{devlin2019bert}.
Given a paragraph $c_{1} \cdots c_{n}$
the output of BERT is a sequence of contextual representation at the character-level:
\begin{equation}
\label{bert}
	\bm{h}_1\cdots \bm{h}_n=\operatorname{BERT}\left(c_1 \cdots c_n\right)
\end{equation}
where $h_1\cdots h_n$ is our desired representation.

\paragraph{Mention Detection}
The CRF decoder is exploited to obtain the sequence labeling outputs.
First, the encoder outputs $h_1\cdots h_n$ are transformed into tag scores at each position:
\begin{equation}
	\bm{o}_i=\bm{W}\bm{h}_i+b
\end{equation}
where $\bm{W} \in \mathcal{R}^{d_h \times |tags|}$, $|tags|$ is the number of tag classes.
Then for each tag sequence $y_{1} \dots y_{n}$, its probability can be computed by:
\begin{equation}
	p (y_{1} \dots y_{n})=  \frac { \exp(\sum_{i=1}\bm{o}_{i, y_i}  +  \bm{T}_{y_i, y_{i-1}} ) } { Z },
\end{equation}
where $\bm{T}$ is a model parameter to indicate output tag transition score and $Z$ is a normalization score.
We can use standard Viterbi to obtain the highest-probability tagging sequence.

\paragraph{Coreference Recognition}
We take the input character representation $h_1\cdots h_n$ obtained by the encoder
as well as the output tag sequence $y_{1} \dots y_{n}$ of mention detection as inputs.
For the output tag sequence, we exploit a simple embedding matrix $\bm{E}$ to convert tags into vectors $\bm{e}_1 \cdots \bm{e}_n$.
Then we concatenate the two kinds of representations,
getting the encoder output $\bm{z}_1 \cdots \bm{z}_n$, where $\bm{z}_i = \bm{h}_i \oplus \bm{e}_i$.
We treat coreference recognition as a binary classification problem.
Given the encoder output $\bm{z}_1 \cdots \bm{z}_n$ and two positions $i$ and $j$,
we judge whether the relation between the two characters $c_i$ and $c_j$ is a coreference or not.
For one fine-grained coreference $c_\text{f}$ and $w_\text{s}$,
we regard all characters in $w_\text{s}$ be connected to $c_\text{f}$.

The score of the binary classification is computed by a simple BiAffine operation,
\begin{equation}
	 \bm{s}_{i,j} = \operatorname{BiAffine}(\bm{z}_i, \bm{z}_j),
\end{equation}
where $\bm{s}_{i,j}$ is one two-dim vector.
We can refer to \cite{DBLP:conf/iclr/DozatM17} for the details of the BiAffine operation,
which has shown strong capabilities in similar tasks \cite{DBLP:conf/acl/ZhangSYXR18}.

\paragraph{Training}
For mention detection, 
the supervised training objective is to minimize the cross-entropy of the gold-standard tagging sequence:
\begin{equation}
	\mathcal{L}_{\text{mention}} = - \log  p (y^g_{1} \dots y^g_{n}),
\end{equation}
where $y^g_{1} \dots y^g_{n}$ is the supervised answer.

For coreference recognition,
we adopt averaged cross-entropy loss overall input pairs as the training objective:
\begin{equation}
	\mathcal{L}_{\text{coref}} = \frac{\sum_{i \in [1,n],j \in [1,n], i \neq j} p (r_{i,j})} { n*(n-1)},
\end{equation}
where $r_{i,j}$ denotes the ground-truth relation.
The probability is computed by:
\begin{equation}
	p (r_{i,j}) = \frac{ \exp(\bm{s}_{i,j}[r_{i,j}]) } { Z_{i,j} },
\end{equation}
where $Z_{i,j}$ is a normalization factor.

We combine losses from the two subtasks together for joint training:
\begin{equation}
	\mathcal{L}_{\text{joint}}=\mathcal{L}_{\text{mention}}+ \alpha \mathcal{L}_{\text{coref}},
\end{equation}
where $\alpha$ is a hyper-parameter to balance the losses of the two subtasks.

\begin{figure}[t]
\centering
\includegraphics[scale=0.5]{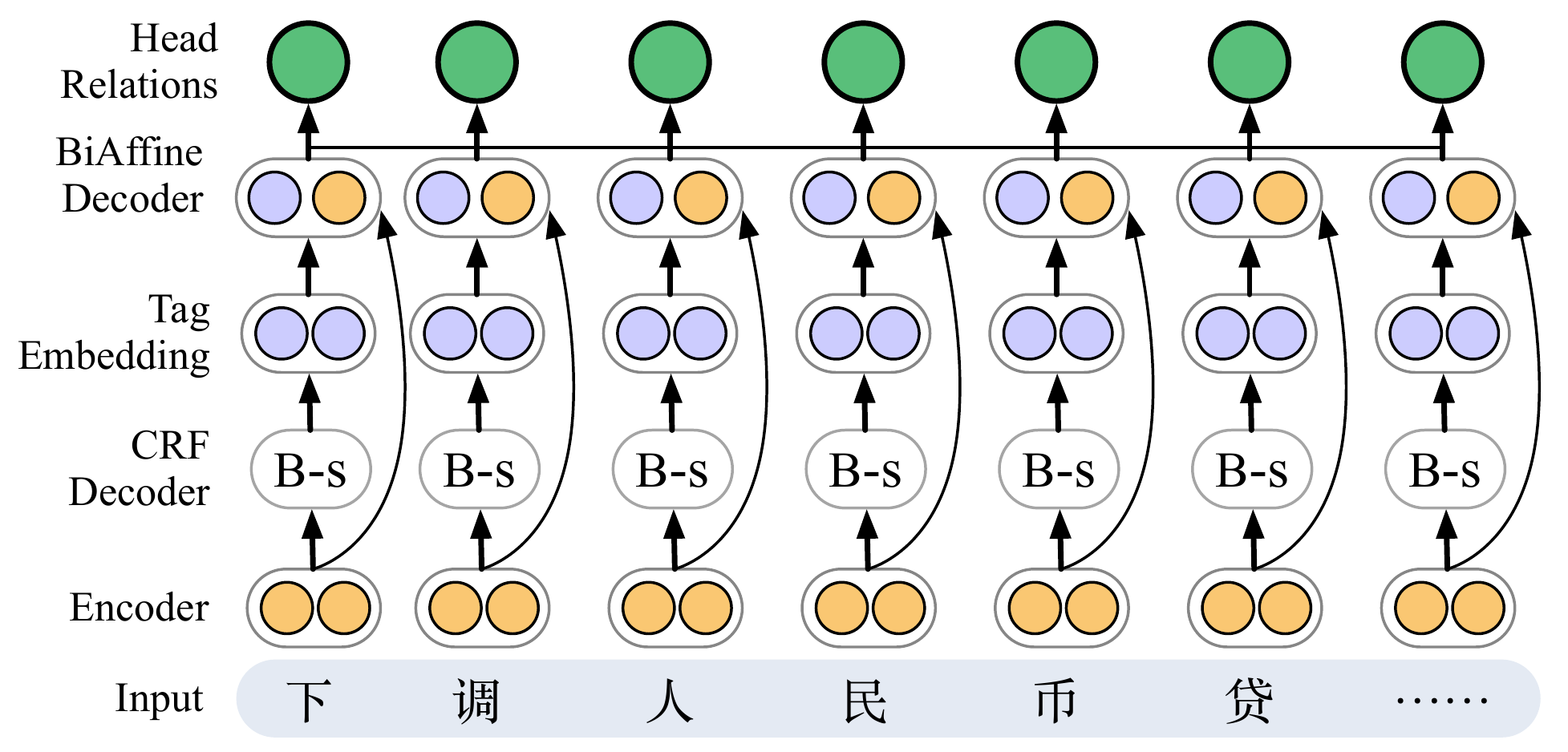}
\caption{Overview of the end-to-end architecture for lexical fusion recognition.}
\label{figjoint}
\end{figure}

\subsection{Sememe-Enhanced Encoder}
\paragraph{Sememes and HowNet}
Sememe is regarded as the minimum semantic unit for the Chinese language,
which has been exploited for several semantic-oriented tasks, such as word sense disambiguation, event extraction and relation extraction
\cite{gu-etal-2018-language,ding-etal-2019-event,li-etal-2019-chinese}.
Our task is also semantic oriented because the lexical fusion and coreference are both semantic related.
Thus sememe should be one kind of useful information for our model.

\begin{figure}[h]
\centering
\includegraphics[scale=0.6]{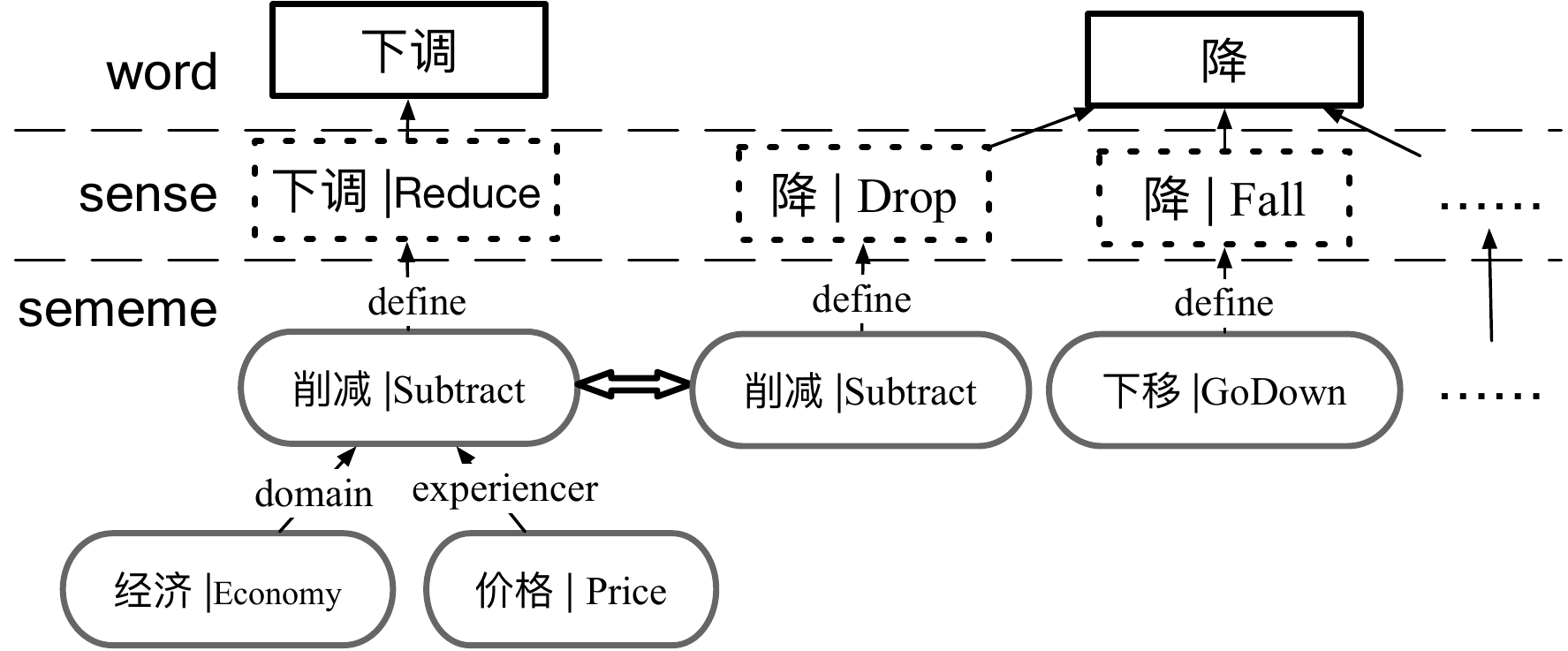}
\caption{The architecture of sememe, sense and word in HowNet, where a sememe may be shared across senses, eg. the ``Subtract''.}
\label{fighownet}
\end{figure}

We follow the majority of the previous work, extracting sememes for characters from HowNet,
a manually-constructed sememe knowledge base.
HowNet defines over 118,000 words (including characters) using about 2,000 sememes.\footnote{Chinese characters are also basic semantic units for meaning expressing like full words.}
Figure \ref{fighownet} illustrates the annotations in HowNet.
We can see that each word is associated with several senses,
and further, each sense is annotated with several sememes,
where sememes are organized by graphs.

\begin{figure}[h]
\centering
\includegraphics[scale=0.5]{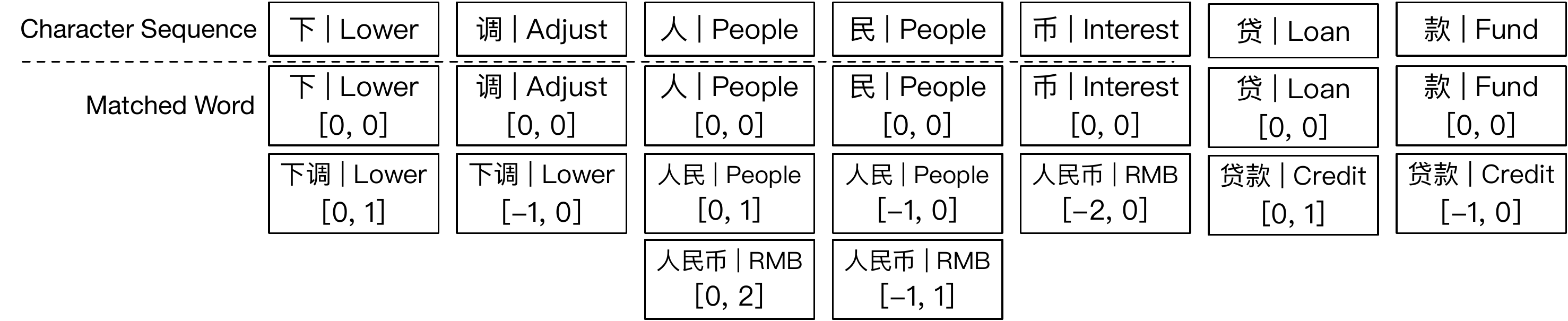}
\caption{\label{char_word}The matched word and relative position for each character in the sequence.}
\end{figure}

\paragraph{Sememe to Character Representation}
For each character $c_i$,
we make use of all possible HowNet words covering $c_i$, as shown in Figure \ref{char_word},
and further extract all the included senses by these words.
Each sense corresponds to one sememe graph, as shown in Figure \ref{fighownet}.
The sememe to character representation is obtained by two steps.
First, we obtain the sense representation by its sememe graph and the position offset of its source word.
Then, we aggregate all sense representations to reach a character-level representation,
resulting in the sememe-enhanced encoder.

We use standard GAT to represent the sememe graph.
Let $\mathrm{sm}_1\cdots \mathrm{sm}_M $ denote all the sememes belonging to a given sense $\mathrm{sn}$,
and their internal graph structure is denoted by $G$,
after apply the GAT module,
we can get:
\begin{equation}
	\bm{h}_{\mathrm{sm}_1} \cdots \bm{h}_{\mathrm{sm}_M}  \text{~~}\mathclap{=}\text{~~} \operatorname{GAT} ( \bm{e}_{\mathrm{sm}_1} \cdots \bm{e}_{\mathrm{sm}_M},  G ),
\end{equation}
where $\bm{e}_{*}$ indicates the embedding operation.

Further, we obtain the first representation part of $\mathrm{sn}$ by averaging over $\bm{h}_{\mathrm{sm}_1} \cdots \bm{h}_{\mathrm{sm}_M}$.
The second part is obtained straightforwardly by the embedding of the position offset of the sense's source word.
The position offset is denoted by $[s,e]$, where $s$ and $e$ indicate the relative position of the start and end characters of the source word to the current character,
which has been illustrated in Figure \ref{char_word}.
We use the position offset as a single unit for embedding.
Following, we concatenate the two parts, resulting in the sense representation:
\begin{equation}
	\bm{h}_{\mathrm{sn}}  = \frac{\sum_{i=1}^m \bm{h}_{\mathrm{sm}_i}}{M} \oplus \bm{e}_{[s,t]} ,
\end{equation}
where $\oplus$ denotes vector concatenation.

Finally, we aggregate all sense representations by global attention \cite{DBLP:conf/emnlp/LuongPM15} with the guide of the BERT outputs to obtain character-level representations.
Let $\{\mathrm{sn}_1, \cdots, \mathrm{sn}_N \}$ denote the set of sense representations for one character $c_i$,
the sememe-enhanced representation for character $c_i$ can be computed by:
\begin{equation}
\begin{split}
    & a_j =\frac{\exp\big( \tanh(\bm{v}[\bm{h}_i\oplus\bm{h}_{\mathrm{sn}_j}])\big) }{\sum_{k=1}^N\exp\big( \tanh(\bm{v}[\bm{h}_i\oplus\bm{h}_{\mathrm{sn}_k}])\big) }, \\
	& \bm{h}_i^{\text{sem}} = \sum_{j=1}^{N} a_j \cdot \bm{h}_{\mathrm{sn}_j},
\end{split}
\end{equation}
where $\bm{v}$ is a model parameter for attention calculation,
and $\bm{h}_1^{\text{sem}} \cdots \bm{h}_n^{\text{sem}}$ are the desired outputs which is used instead of the BERT outputs $\bm{h}_1 \cdots \bm{h}_n$ for decoder.

\section{Experiments}

\subsection{Dataset}
\label{corpus}
\paragraph{Test Data}
We build a lexical fusion dataset manually in this work,
where the raw corpus is collected from SogouCA \cite{Wang2008Automatic},
a news corpus of 18 channels,
including domestic, international, sports, social, entertainment, etc.
The BRAT Rapid Annotation Tool \cite{stenetorp2012brat} is used for annotation.
We label the boundary of mentions in the paragraph,
determine their categories (fusion or separation words),
link characters in the fusion words to the referred separation words,
and finally format the annotation results as triples.

We annotate 17,000 paragraphs by five annotators who major in Chinese linguistics.
After removing the paragraphs without lexical fusion,
the five annotators check the annotations once again to ensure the quality.
Finally, 7,271 lexical fusion cases are obtained with 91\% consensus.

\paragraph{Pseudo Training Dataset}
We construct a pseudo lexical fusion corpus to train the models,
by making use of an online lexicon where words are offered with explanatory notes.\footnote{https://cidian.911cha.com/}
For each two-character word, we check if it can be split into two separation words.
If successful, we obtain one context-independent triple.
Finally, we collect 1,608 well-formed triples of lexical fusions
and treat them as seeds to construct pseudo training instances.
Note that the fusion words of these triples are currently acceptable and widely used by users,
such as ``停车(parking vehicles)$\leftrightarrow$停放(parking)/车辆(vehicle)''.
Then, we search for the paragraphs containing all three words of a certain triple,
regarding them as valid cases of Chinese lexical fusion\cite{mintz2009distant}.
Finally, we obtain 11,034 paragraphs,
which are divided into training and development sections
for model parameter learning and hyper-parameter tuning, respectively.

Table \ref{dataset} summarizes the overall data statistics of the training,
development and testing sections,
including the number of cases (lexical fusion),
the averaged paragraph length (by character count) and the number of unique triples, respectively.

\setlength{\tabcolsep}{9pt}
\begin{table}[h]
\centering
\begin{tabular}{l|ccccc}
\toprule  &  \#Case   & \#U.Triple  & \#A.Length & \#N.Case  & \#N.U.Triple  \\ \midrule
Training & 7,400 & 1,531 & 106 & - & - \\
Development & 2,213 & 1,001 & 105 & 94 & 77 \\
Testing & 7,271 & 1,661 & 91 & 3959 & 957 \\
\bottomrule
\end{tabular}
\caption{\label{dataset} Data statistics of our corpus, where A. , U. and N. indicate average, unique and new, respectively.}
\end{table}

\subsection{Evaluation}
A triple is regarded as correct if all the three elements are correctly recognized
and meanwhile on their exact positions,
We calculate triple-level and fine-grained precision (P), recall (R) and F-measure (F) values,
and adopt the triple-level values as the major metrics to evaluate the model performance.
We also calculate mention-level P, R and F values to evaluate the performance of mention detection.

\subsection{Settings}
All the hyper-parameters are determined according to the performance on the development dataset.
We exploit the pretrained basic BERT representations as inputs for encoder \cite{devlin2019bert}.\footnote{https://github.com/google-research/bert}
The tag, sememe and position embedding sizes are set to 25, 200 and 12, respectively.
The head number of GAT is 3.
We exploit dropout with a ration of 0.2 on the encoder outputs to avoid overfitting \cite{DBLP:journals/jmlr/SrivastavaHKSS14}, 
and optimize all model parameters (including the BERT part)
by standard back-propagation using Adam \cite{kingma2014adam}
with a learning rate $0.001$.

\setlength{\tabcolsep}{2pt}
\begin{table}[h]
\centering
\begin{tabular}{l|ccc|ccc|ccc|ccc}
\toprule
\multirow{3}{*}{~~~~~~~~~~Model} & \multicolumn{6}{c|}{Entity Detection} & \multicolumn{6}{c}{Coreference Recognition}  \\ \cline{2-13}
  & \multicolumn{3}{c|}{Fusion} & \multicolumn{3}{c|}{Separation} & \multicolumn{3}{c|}{Fine-Grained} & \multicolumn{3}{c}{Triple-Level} \\ \cline{2-13}
\multirow{3}{*}{} & \multicolumn{6}{c|}{Entity Detection} & \multicolumn{6}{c}{Coreference Recognition}  \\ \cline{2-13}
 & P & R & F & P & R & F & P & R & F & P & R & F \\
\hline 
\hline
\multicolumn{13}{c}{\texttt Pipeline} \\ \hline
\texttt{basic} & 83.10 & 83.88 & 83.49 & 82.84 & 84.46 & 83.64 & 76.09 & \textbf{76.66} & 76.37 & 74.26 & \textbf{75.23} & 74.74 \\
\texttt{GAT\small{(\text{word},\text{Hownet})}} & 84.57 & 83.02 & 83.79 & 86.92 & 81.79 & 84.28  & 85.07 & 74.27 & 79.30  & 85.02 & 73.27 & 78.71 \\
\hline\hline
\multicolumn{13}{c}{\texttt Joint} \\ \hline
\texttt{basic} & 82.85 & 81.40 & 82.12  & 85.79 & 82.46 & 84.09  & 81.65 & 74.32 & 77.81  & 80.32 & 72.94 & 76.45  \\
\texttt{GAT\small{(\text{char},\text{pseudo})}} & 84.61 & 83.15 & 83.87  & 86.83 & 82.58 & 84.56 & 83.19 & 75.81 & 79.33 & 81.57 & 74.46 & 77.85\\
\texttt{GAT\small{(\text{char},\text{Hownet})}} & 84.79 & \textbf{84.27} & 84.53 & 88.07 & 81.75 & 84.79 & 84.71 & 75.15 & 79.64 & 83.72 & 73.87 & 78.49\\
\texttt{GAT\small{(\text{word},\text{pseudo})}} & 84.55 & 83.69 & 84.12  & 85.70 & \textbf{84.82} & 85.26  & \textbf{87.88} & 73.12 & 79.82  & \textbf{89.38} & 70.62 & 78.90  \\
\texttt{GAT\small{(\text{word},\text{Hownet})}} & \textbf{85.18} & 84.16 & \textbf{84.67} & \textbf{88.64} & 83.98 & \textbf{86.25} & 85.63 & 75.60 & \textbf{80.30} & 86.19 & 74.02 & \textbf{79.64}\\
\bottomrule
\end{tabular}
\caption{\label{ex-result} Final results on the test dataset, where \texttt{basic} indicates no sememe information is used.}
\end{table}

\subsection{Results}
Table \ref{ex-result} shows the final results on the test dataset.
Aiming to investigate the influence of sememe structure in detail,
we construct a pseudo graph structure $G'$ for comparison where all sememes inside a sense are mutually connected.
In addition, we conduct comparisons by using only the sememes from characters (single-character words),
in order to explore the effect of the word-level sememe information.
Our final model, the end-to-end model with word-level sememe-enhanced encoder (i.e., \texttt{joint + GAT\small{(\text{word}\_\text{real})}}),
achieves competitive performance, where the triple-level F-measure reaches $79.64$,
which is the best-performance model, significantly better than the basic model without using the sememe information.\footnote{The p-value is below 0.0001 by using pairwise t-test.} 
In addition to the contents of \ref{ex-result}, 
we also tried the pipeline approach as a contrast.
The triple-level F-measure of pipeline models with or without word-level sememe graph are 78.71\% and 74.74\%, respectively.

%

According to the results,
our proposed sememe-enhanced encoder is effective, bringing significant improvements over the corresponding basic model.
The improvements on the triple-level F-measure is $78.45-76.45=2.04$.
The performance differences of our sememe-enhanced encoder by using the pseudo sememe graph and the real sememe graph indicate the effectiveness of sememe structure.
The real sememe graph can give consistently better performance (i.e., an increase of $0.69$ on average) than the pseudo graph.
As shown, the word-level sememe structure can obtain F-measure improvements of $79.64-78.49=1.15$,
indicating the usefulness of the word-level information.


\subsection{Analysis}
\paragraph{Influence of Lexical Fusion Types}
We investigate the model performance with respect to different lexical fusion types.
We classify the fusion word types by IV/OOV according to the training corpus,
and further differentiate a fine-grained coreference by whether the fusion character is borrowed from its separation word (denoted by A) or not (denoted by B).
Table \ref{recall} shows the results.
Our models perform better for the IV categories than the OOV, which confirms with our intuition.
In addition, we divide the IV/OOV further into AA and AB categories.\footnote{We ignore the BB categories as the number is below 10. }
We can find that AB is much more difficult,
obtaining only 41.1 of the F1 score.
Further, by examining the overall performance of fine-grained coreference of type A and B,
we can see that Type B leads to the low performance mainly.
The performance gap between the two types is close to 50 on the F1 score.


\setlength{\tabcolsep}{7pt}
\begin{table}[h]
\centering
\begin{tabular}{l|rr|rr|rr}
\toprule
\multirow{2}{*}{Model} & \multicolumn{2}{c|}{IV } & \multicolumn{2}{c|}{OOV } & \multicolumn{2}{c}{Fine-Grained} \\ \cline{2-7}
 & AB & AA &  AB & AA & B & A\\
\midrule
Joint & 86.4 & 94.5 & 30.4 & 81.8 & 28.1 & 82.4\\
\texttt{GAT\small{(\text{word}\_\text{real})}} & \textbf{89.5} & \textbf{94.7} &  \textbf{41.1} & \textbf{82.9} & \textbf{36.5} & \textbf{83.7}\\
\bottomrule
\end{tabular}
\caption{\label{recall}F1 scores by different lexical fusion types. }
\end{table}

\begin{figure}[h]
\centering
\includegraphics[scale=0.47]{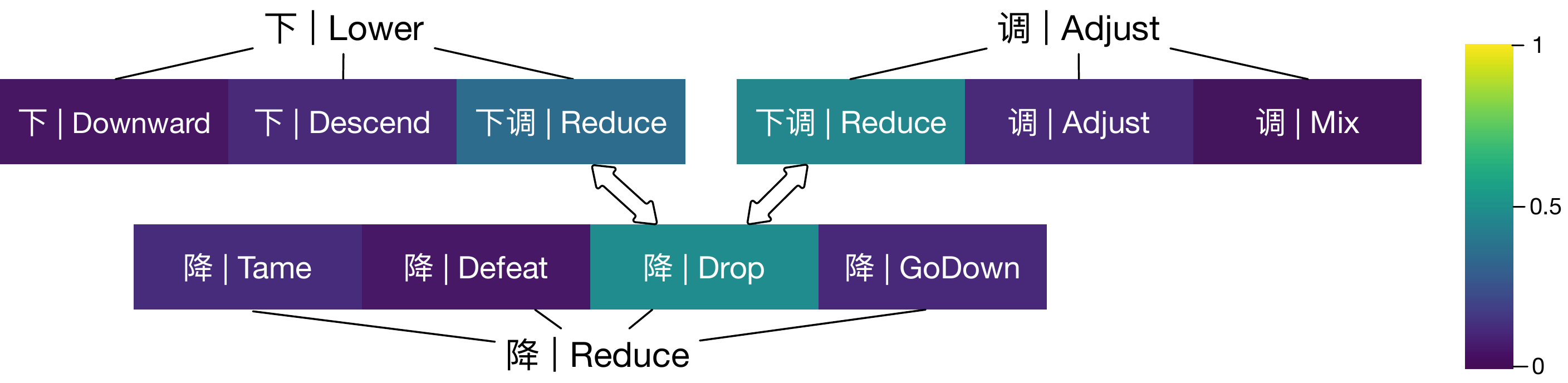}
\caption{The sense-level attention weights of ``降$\leftrightarrow$下调 (reduce)'', where the colored squares represent senses, the senses with shared sememes are linked with double arrows.}
\label{snsattn}
\end{figure}

\paragraph{Effect of Sememe Information}
In order to understand the sememe-enhanced encoder in-depth,
we examine the sense-level attention weights by an example.
As shown in Figure \ref{snsattn},
a fine-grained coreference ``降$\leftrightarrow$下调 (reduce)'' is used for illustration,
where the three characters ``下''(lower),  ``调''(adjust) and ``降''(reduce) are studied.
Each character includes a set of senses, which are listed by the squares.\footnote{Senses with very low weights are filtered.}
We can see that senses with shared sememes can mutually enhance their respective attention weights,
which is consistent with the goal of coreference recognition.
It is difficult to establish such a connection without using sememes.

\begin{figure}[htb]
\centering
\subfigure[\label{coref_type}The triple-level F1 scores in terms of mention order.]{
\begin{minipage}[t]{0.5\linewidth}
\centering
\includegraphics[scale=0.86]{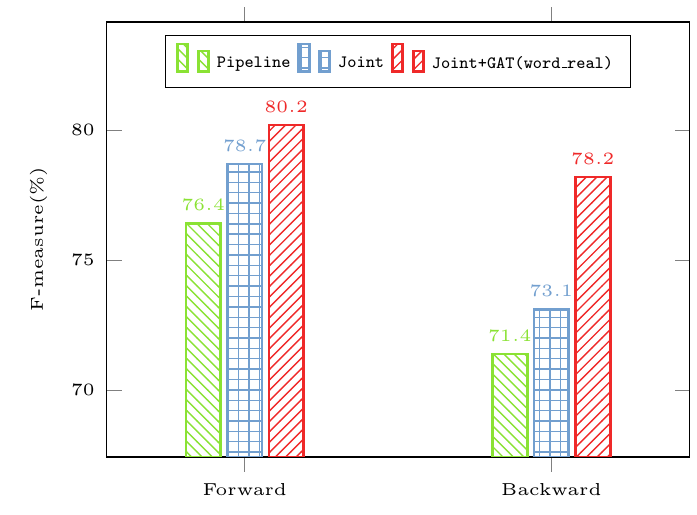}
\end{minipage}%
}%
\subfigure[\label{distance}The triple-level F1 scores with respect to the distance between the separation words.]{
\begin{minipage}[t]{0.5\linewidth}
\centering
\includegraphics[scale=0.8]{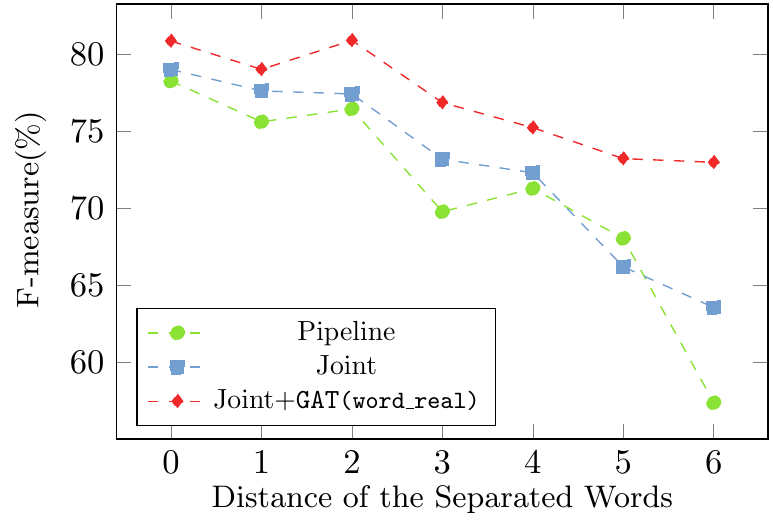}
\end{minipage}%
}%

\centering
\caption{Impact of the order and the distance.}
\end{figure}

\paragraph{Impact of Order of Mentions}
Figure \ref{coref_type} shows the F-measure values according to the relative order
of the mentions.
Intuitively,
the recognition of the forward references is more difficult
than that of the backward references.
The results confirm our intuition,
where the F1 score of backward reference is 4.2 points higher on average.
In addition, our final model can improve performance of both types significantly.

%
%

\paragraph{Impact of the Separation Word Distance}
The distance between the two separation words should be an important factor in coreference recognition.
Intuitively, as the distance increases, the difficulty should be also increased greatly.
Here we conduct analysis to verify this intuition.
Figure \ref{distance} shows the comparison results,
which is consistent with our supposition.
In addition, we can see that our final end-to-end model behaves much better,
with relatively smaller decreases as the distance increases.

\section{Related Work}
Coreference recognition has been investigated extensively in NLP for decades \cite{mccarthy1995using,cardie1999noun,ng-cardie-2002-improving,elango2005coreference}.
Lexical fusion can be regarded as one kind of coreference,
however, it has received little attention in the NLP community.
Our proposed models are inspired by the work of neural coreference resolution \cite{fern2016detection,clark-manning-2016-improving,xu-etal-2017-local,lee-etal-2017-end,DBLP:conf/acl/ZhangSYXR18}.
We adapt these models by considering task-specific features of Chinese lexical fusion,
for example, enhancing the encoder with a GAT module for structural sememe information.

Another most closely-related topic is abbreviation \cite{Zhong:1985}.
There have been several studies on abbreviation prediction,
recovery and dictionary construction \cite{Xu:2008,li-yarowsky-2008-unsupervised,Liu:2009,Zhang:2017}.
Lexical fusion is different from abbreviation in many points.
For example, abbreviation always refers to one inseparable mention,
which is not necessary for lexical fusion.
Besides, lexical fusion should abide by certain word construction rules,
while abbreviation is for free.

BERT and its variations have achieved the leading performances for GLUE benchmark datasets \cite{devlin2019bert,liu-etal-2019-multi,liu2020roberta}.
For the close tasks such as coreference resolution and relation extraction,
BERT representations have also shown competitive results \cite{joshi-etal-2019-bert,lin2019bert}, 
which inspires our work by using it as basic inputs aiming for competitive performance.

The sense and sememe information has been demonstrated effective
for several semantic-oriented NLP tasks \cite{niu2017improved,gu-etal-2018-language,DBLP:conf/aaai/ZengYT0S18,ding-etal-2019-event,qi-etal-2019-modeling}.
HowNet offers a large knowledge base of sememe-based \cite{dong2003hownet},
which has been adopted for sememe extraction.
We encode the sememes by the form of a graph naturally,
and then exploit GAT to enhance our task encoder.

\section{Conclusion}
In this work, we introduced the task of lexical fusion recognition in Chinese 
and then presented an end-to-end model for the new task.
BERT representation was exploited as the basic input for the models,
and the model is further enhanced with the sememe knowledge from HowNet by graph attention networks.
We manually annotated a benchmark dataset for the task,
which was used to evaluate the models.
Experimental results on the annotated dataset
indicate the competitive performance of our final model,
and the effectiveness of the joint modeling and the sememe-enhanced encoder.
Analysis is offered to understand the new task and the proposed model in-depth.

\bibliographystyle{coling}
\bibliography{coling2020}

\end{CJK}
\end{document}